\documentclass[runningheads]{llncs}
\usepackage{graphicx}
\usepackage{booktabs} 
\usepackage{float}
\usepackage{algorithm,algorithmic}
\usepackage{bbm}
\usepackage{amssymb}
\usepackage{diagbox}
\graphicspath{{media/}}
\usepackage{float}
\usepackage{amsmath}
\usepackage{amsfonts}
\usepackage{multirow}
\usepackage{multicol}
\usepackage{hyperref}
\hypersetup{colorlinks=true,
            linkcolor=blue,
            anchorcolor=blue,
            citecolor=blue,
            urlcolor=blue,
            }
\usepackage{bbding}
\begin{document}
%
\title{Self-Paced Sample Selection for Barely-Supervised Medical Image Segmentation}
\titlerunning{Self-Paced Sample Selection}
%


%

\author{Junming Su\inst{1,2,\dag} \and
Zhiqiang Shen\inst{1,2,\dag} \and
Peng Cao\inst{1,2,3(}\Envelope\inst{)} \and
Jinzhu Yang\inst{1,2,3} \and
Osmar R. Zaiane\inst{4}}
\authorrunning{J. Su et al.}
%
\institute{Computer Science and Engineering, Northeastern University, Shenyang, China \and
Key Laboratory of Intelligent Computing in Medical Image of Ministry of Education, Northeastern University, Shenyang, China \and
National Frontiers Science Center for Industrial Intelligence and Systems Optimization, Shenyang, China \\
\email{caopeng@mail.neu.edu.cn} \and
Alberta Machine Intelligence Institute, University of Alberta, Edmonton, Canada}

\maketitle              

\renewcommand{\thefootnote}{}
\footnotetext{\dag \ Junming Su and Zhiqiang Shen contributed equally to this work.}
\begin{abstract}
The existing barely-supervised medical image segmentation (BSS) methods, adopting a registration-segmentation paradigm, aim to learn from data with very few annotations to mitigate the extreme label scarcity problem.
However, this paradigm poses a challenge: pseudo-labels generated by image registration come with significant noise. 
To address this issue, we propose a self-paced sample selection framework (SPSS) for BSS. 
Specifically, SPSS comprises two main components: 1) self-paced uncertainty sample selection (SU) for explicitly improving the quality of pseudo labels in the image space, and 2) self-paced bidirectional feature contrastive learning (SC) for implicitly improving the quality of pseudo labels through enhancing the separability between class semantics in the feature space. 
Both SU and SC are trained collaboratively in a self-paced learning manner, ensuring that SPSS can learn from high-quality pseudo labels for BSS. 
Extensive experiments on two public medical image segmentation datasets demonstrate the effectiveness and superiority of SPSS over the state-of-the-art.
Our code is release at \textit{https://github.com/SuuuJM/SPSS}.

\keywords{Barely-Supervised Learning \and Self-Paced Learning \and Contrast Learning.}
\end{abstract}

\section{Introduction}
\begin{figure}[!t]
    \centering
    \includegraphics[width=\textwidth]{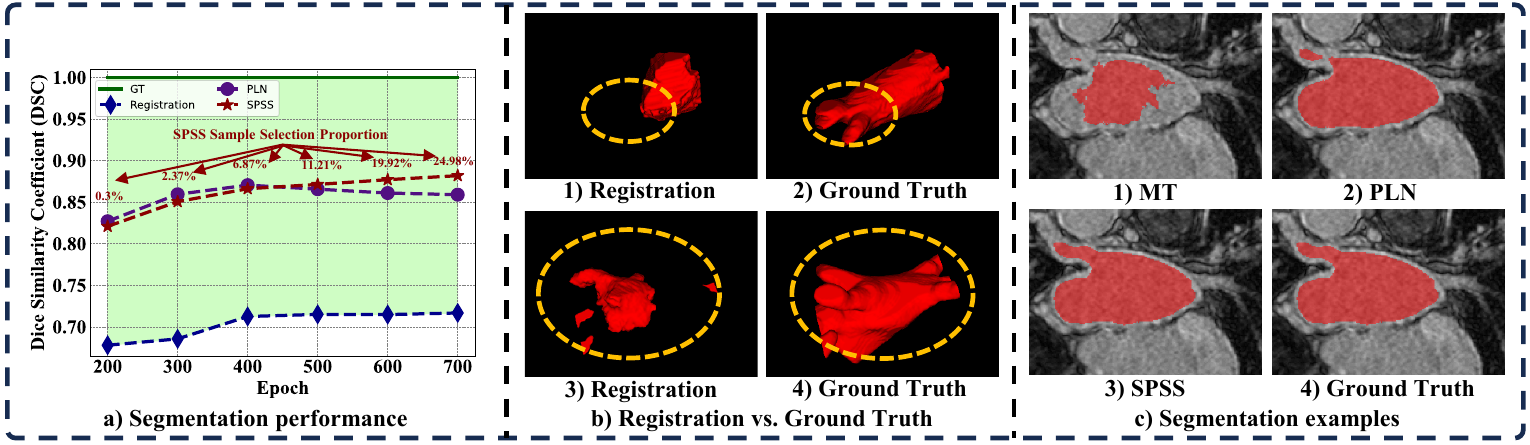}
    \caption{Illustration of a) segmentation performance of PLN and SPSS, b) registration noise, and c) qualitative results of MT, PLN, and SPSS. Note that the green region in a) is the performance drop from ground truth caused by registration noise.}
    \label{introduction_image}
\end{figure}

With the continuous development of deep learning-based methods, medical image segmentation has been significantly advanced \cite{chen2018encoder,ronneberger2015u}, relying on large amounts of labeled data. However, annotating medical images at the pixel level is laborious and requires professional knowledge, making large-scale labeled data expensive or even unavailable.
Semi-supervised learning (SSL) \cite{van2020survey} primarily builds upon the pseudo labeling \cite{lee2013pseudo} and consistency regularization \cite{tarvainen2017mean} techniques, achieving state-of-the-art (SOTA) performance on common label scarcity scenarios, e.g., with 10\% labeled data, for medical image segmentation \cite{yu2019uncertainty}. However, this line of methods faces a critical limitation when encountering the extreme label scarcity problem where the training set includes a barely-annotated labeled set with single-slice annotations and an unlabeled set with numerous images. As depicted in Fig. \ref{introduction_image}c.1), for instance, the representative SSL method, Mean-Teacher \cite{tarvainen2017mean}, yields unsatisfactory performance on the 3D left atrium segmentation task \cite{xiong2021global} with barely-annotated data.

Barely-supervised learning (BSL) has emerged to solve such extreme label scarcity for further alleviating the annotation burden \cite{li2022pln,cai2023orthogonal,cai20233d}. 
SOTA barely-supervised medical image segmentation (BSS) methods, such as PLN \cite{li2022pln}, adopt a registration-segmentation paradigm to construct 3D pseudo-labels from single-slice annotations.
This paradigm mainly includes the following procedure:
1) a registration module constructs volumetric pseudo labels using single-slice annotations of barely labeled data, 2) a teacher segmentation model generates pseudo labels for both barely labeled data and unlabeled data, and 3) a student segmentation model is trained based on the fused pseudo labels produced by the first two steps.
However, the pseudo labels generated by image registration are unreliable and noisy.
This registration noise inevitably degrades the performance of the segmentation module, resulting in inferior performance. 
To illustrate it, we conducted a preliminary experiment on the left atrial segmentation task for BSS.
Quantitatively, as depicted in Fig. \ref{introduction_image}a), the registration performance only reaches 60\%-70\% in terms of dice similarity coefficient (DSC), indicating a significant gap between the registered 3D pseudo labels and the ground truth annotations. Qualitatively, the pseudo labels also differ substantially from ground truth [Fig. \ref{introduction_image}b)]. 
Based on the registration-segmentation paradigm, the state-of-the-art BSS approach, PLN \cite{li2022pln}, shows a performance drop in the latter epochs of the training stage [Fig. \ref{introduction_image}a)] and fails in the challenging boundary regions [Fig. \ref{introduction_image}c.2)].
\textit{A core problem in the registration-segmentation paradigm naturally arises: how to improve the quality of pseudo labels during training for barely-supervised medical image segmentation.}

To this end, we propose a self-paced sample selection framework (SPSS), tailored for BSS. 
As shown in Fig. \ref{introduction_image}a), the DSC curve of our SPSS is higher than that of PLN \cite{li2022pln} during the latter epochs of the training phase, suggesting that the segmentation performance of SPSS is consistently improved due to the selected high-quality pseudo labels.
Considering the difficulty in directly enhancing the capability of the registration module with a limited number of labeled slices\cite{li2022pln}, we provide an alternative that aims at strengthening the segmentation capability to alleviate the impact of registration noise by improving the quality of pseudo labels.
\textit{The main idea of SPSS lies in improving the quality of pseudo labels in both the image and feature spaces by self-paced sample selection\footnote{Samples refer to pixels in both the image and feature spaces.}.}
To realize this idea, SPSS consists of two components: 1) self-paced uncertainty sample selection (SU), and 2) self-paced bidirectional feature contrastive learning (SC). 
On the one hand, SU explicitly selects high-quality pseudo labels in the image space at the voxel level. 
On the other hand, SC enhances the separability of class semantics in the feature space through bidirectional feature contrastive learning. 
Both SU and SC are trained collaboratively in a self-paced learning manner, guaranteeing that SPSS can learn from high-quality pseudo labels for BSS. 
We evaluate the proposed method on two benchmark datasets for BSS. Experimental results show that our SPSS has significant improvements over the state-of-the-art. 
For instance, compared with the state-of-the-art BSS method, i.e. PLN \cite{li2022pln}, our SPSS obtains DSC gains of 2.18\% and 1.25\%; it also outperforms the representative semi-supervised segmentation method UA-MT \cite{yu2019uncertainty} by a large margin of 23.08\% and 13.98\% in terms of DSC on the LA and KiTS datasets with 20\% barely-labeled data, respectively.

In a nutshell, our contributions mainly include:
\begin{itemize}
    \item We design SU and SC to effectively improve the quality of pseudo labels in both the image and feature spaces.
    \item We further train SU and SC collaboratively through a self-paced learning scheme, to construct a novel BSS framework SPSS.
    \item Extensive experiments on left atrial and kidney segmentation tasks validate the effectiveness of SPSS and suggest that our SPSS provides an alternative aiming to strengthen segmentation capability, alleviating the impact of registration noise by improving the quality of pseudo-labels.
\end{itemize}



\section{Methodology}
The training set $\mathcal{D}=\{\mathcal{D}_L,\mathcal{D}_U\}$ contains a labeled dataset $\mathcal{D}_L=\{X^l_i,y^l_{k_i}\}_{i =1,..., N_L}$ and an unlabeled dataset $\mathcal{D}_U=\{X^u_j\}_{j=1,..., N_U}$, where $X^l_i$ represents the $i_{th}$ labeled image with the corresponding single-slice annotation $y^{k_i}$ (only the $k_{th}$ slice in $X^l_i$ has annotation), $X^u_j$ denotes the $i_{th}$ unlabeled image.
\subsection{Overview}
\begin{figure}[!t]
    \centering
    \includegraphics[width=12cm]{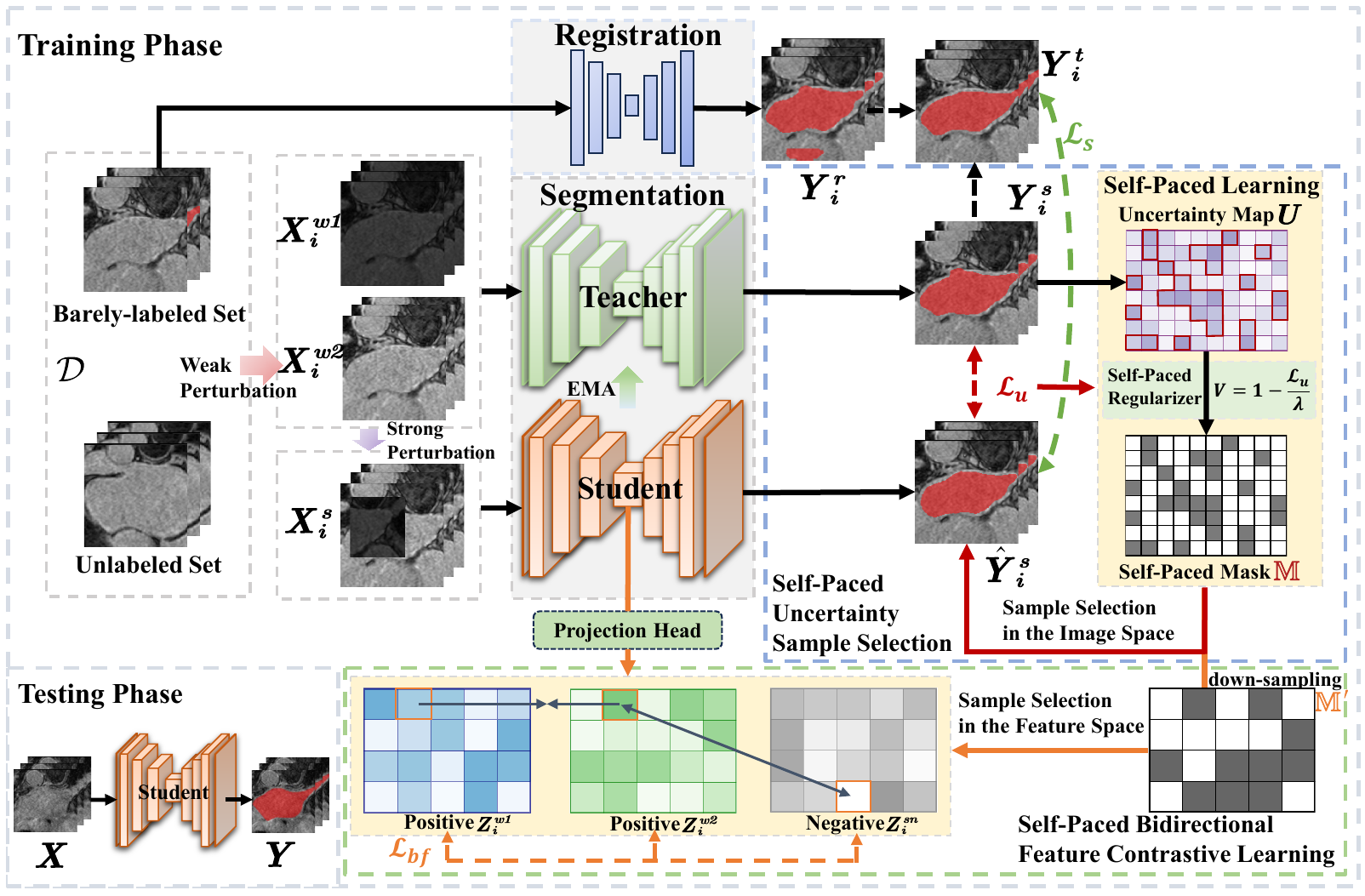}
    \caption{Overview of the proposed self-paced sample selection framework (SPSS). 
    SPSS includes: 1) a self-paced uncertainty sample selection strategy (SU) for explicitly pseudo labels selection in the image space and 2) a self-paced bidirectional feature contrastive learning scheme (SC) for class semantics discrimination in the feature space.
    }
    \label{frame}
\end{figure}

To address the issue of the noisy pseudo labels generated by registration model in the registration-segmentation paradigm, we propose a self-paced sample selection framework (SPSS) [Fig. \ref{frame}] to improve the quality of pseudo labels in both the image and feature spaces.
Specifically, SPSS involves 1) a self-paced uncertainty sample selection strategy (SU) to explicitly select high-quality pseudo-labels in the image space and 2) a self-paced bidirectional feature contrastive learning scheme (SC) to enhance the discrimination of class semantics in the feature space. SU and SC are trained collaboratively in a self-paced learning manner.

\subsection{Self-Paced Uncertainty Sample Selection}

Thresholding is a common strategy for pseudo label selection \cite{sohn2020fixmatch}.
A fixed threshold is arbitrary for pixel/voxel-level sample selection of an image with numerous pixels. However, it is challenging to determine the appropriate thresholds manually. 
Motivated by the concept of self-paced learning \cite{meng2017theoretical,shujun2020}, instead of setting the pixel-level thresholds manually, we propose SU to gradually select high-quality pseudo labels from easy to hard at the voxel level. 
During the self-paced learning, a course of difficulty ranking and an age parameter $\lambda$ are required.
In this study, we determine the self-paced course and the age parameters $\lambda$ based on the model's learning status by considering two aspects: model uncertainty and loss.
Formally, SU calculates a self-paced mask $\mathbbm{M}$ to select $K$ most certain voxels with high-quality pseudo labels in the current iteration. The self-paced mask $\mathbbm{M}$ is defined as: 
\begin{equation}
\mathbbm{M} = \mathbb{I}\{U < Sort(U)[K]\}, \label{M}
\end{equation}
where $\mathbb{I}\{\cdot\}$ represents the indicator function, $Sort(\cdot)$ is an ascending sorting function, 
$K$ indicates the index of the $K_{th}$ most certain voxel, 
and $U$ denotes the uncertainty map of the model for the current input. 
We employ Monte Carlo dropout \cite{gal2016dropout} to estimate the uncertainty $U$: $U^{\prime}=\frac{1}{T}\sum\limits_{t}p_t^c \quad \text{and} \quad U=-\sum\limits_{c}U^{\prime}logU^{\prime}$, where $p_t^c$ is the probability of the $c_{th}$ class in the $t_{th}$ prediction.
The number of certain voxels $K$ is determined by:
$K=R_{conf}\cdot \mathcal{H} \cdot \mathcal{W} \cdot \mathcal{D}$, 
where $\mathcal{H}$, $\mathcal{W}$, and $\mathcal{D}$ represent the height, width, and depth of the input image, respectively. $R_{conf}$ denotes the confident ratio depending on the model's status:
\begin{equation}
    R_{conf}=
    \left\{
    \begin{array}{ll} 
        0.1*min(\xi(t)\cdot\tau,1), & \text{if } \mathcal{L}_u\geq \lambda \\
        v\cdot min( \xi(t)\cdot\tau,1), & \text{otherwise } \mathcal{L}_u < \lambda
    \end{array}
    \right.
    \label{Rconf}
\end{equation}
where $\tau$, $\xi(t)$, and $\lambda$ are the temperature coefficient, the warm-up function, and the age parameters, respectively. $\mathcal{L}_{u}$ refers to the unsupervised loss that indicates the model's status. We define the self-paced weight $v$ as:
$v=(1-\frac{\mathcal{L}_{u}}{\lambda})$.
In the initial training stage, i.e., $\mathcal{L}_u\geq \lambda$, the model may consider all voxels as hard samples, leading to a scenario where no samples are selected. To address this, we introduce a warm-up in the early stage of self-paced learning. 
The warm-up function $\xi(t)$ increases as the number of training iterations increases, controlling the confident proportion $R_{conf}$ for gradually including samples during training. 
As training progresses, since the model becomes more reliable, i.e., $\mathcal{L}_u < \lambda$, more high-quality samples can be involved for training. 

\subsection{Self-Paced Bidirectional Feature Contrastive Learning}
Enhancing the separability between different class semantics in the feature space can further improve the quality of pseudo labels \cite{chen2020simple,zhao2023rcps}.
Nevertheless, it is crucial to consider how to select appropriate positive and negative pixels in the feature space while avoiding false negatives.
To this end, we introduce a novel self-paced bidirectional feature contrastive learning scheme (SC), where the contrastive samples are selected in a self-paced learning manner [Fig. \ref{frame}].

Specifically, SC utilizes the self-paced mask $\mathbbm{M}$ generated in SU to screen positive and negative samples for contrastive learning. 
We consider the corresponding positions in the two weakly perturbed feature maps $Z_i^{w1}$ and $Z_i^{w2}$ that have consistent predictions as positive samples.
Meanwhile, the samples in the strong-perturbed feature map $Z_i^{sn}$ with different predictions from the positive samples are regarded as negative ones. Due to the significantly larger number of negative samples compared to positive samples, we further select the top $K$ most confident negative samples to prevent an overwhelming imbalance of negatives.
Based on the selected positive and negative samples, we define the bidirectional feature contrast loss $\mathcal{L}_{bf}$ as:
\begin{equation}
\mathcal{L}_{bf} = \sum\limits_{z_i^{w1}\in Z_i^{w1},z_i^{w2}\in Z_i^{w2},z_i^{sn}\in Z_i^{sn}}\mathcal{L}_{f}(z_i^{w1},z_i^{w2},z_i^{sn})+\mathcal{L}_{f}(z_i^{w2},z_i^{w1},z_i^{sn}) \label{num1}
\end{equation}
where 
$Z_i^{w[1/2]}$ and $Z_i^{sn}$ are the selected positive and negative feature maps. These feature maps are obtained by multiplying the feature maps $F_i^{w1/2}$ and $F_i^{sn}$ from the projection head with the self-paced mask $\mathbbm{M}$.
Note that we down-sample $\mathbbm{M}$ to match the size of the feature maps. 
$z_i^{w1}$, $z_i^{w2}$, $z_i^{sn}$ are the corresponding samples in these feature maps.
$\mathcal{L}_{f}$ is the feature contrast loss, defined in Eq. \ref{L_f}, for pulling together the positive samples and pushing away the negative samples.
\begin{equation}
\mathcal{L}_f(z_i^{w1},z_i^{w2})=-log\frac{exp\{cos(z_i^{w1},z_i^{w2})/\tau\}}{exp\{cos(z_i^{w1},z_i^{w2})/\tau\}+\sum\limits_{j=1}^{K}exp\{cos(z_i^{w1},z_{ij}^{sn})/\tau\}} \label{L_f}
\end{equation}
where $\mathrm{cos}(\cdot)$ is the cosine similarity function.

\subsection{Loss Function}
The overall loss for training SPSS is defined as: $\mathcal{L}=\mathcal{L}_{s}+\mathcal{L}_{u}+\mathcal{L}_{bf}$,
where $\mathcal{L}_{bf}$, $\mathcal{L}_s$, and $\mathcal{L}_u$ represent the aforementioned contrastive loss, the supervised loss, and the unsupervised loss, respectively.
The supervised loss $\mathcal{L}_s$ is calculated by:
$\mathcal{L}_{s}=\mathcal{L}_{Dice}(\hat{Y}_i^s,Y_i^t)+\mathcal{L}_{CE}(\hat{Y}_i^s,Y_i^t)$, where $Y_i^t$ is the weighted fusion of the registration pseudo labels $Y_i^r$ and the segmentation pseudo labels $Y_i^s$.
Based on the self-paced mask $\mathbbm{M}$, the unsupervised loss $\mathcal{L}_u$ is formulated as:
\begin{equation}
\mathcal{L}_{u}=\mathcal{L}_{Dice}(\mathbbm{M} \cdot \hat{Y}^s_i, \mathbbm{M} \cdot Y^s_i)+\mathcal{L}_{CE}(\mathbbm{M} \cdot \hat{Y}^s_i, \mathbbm{M} \cdot Y^s_i)
\end{equation}
where $X^s_i$, $Y^s_i$, and $\hat{Y}^s_i$ denote the strong-perturbed image, the corresponding pseudo label, and the prediction of the student model, respectively.
We apply the CutMix operation \cite{yun2019cutmix} on the two weak-perturbed images $X^{w1}_i$ and $X^{w2}_i$, as well as their corresponding pseudo labels $Y^{w1}_i$ and $Y^{w2}_i$, to generate $X^{s}_i$ and $Y^{s}_i$.

\section{Experiments and Results}
\label{section4}

\textbf{Dataset} We evaluate the proposed SPSS on the 2018 Left Atrial Segmentation Challenge (LA) dataset \cite{xiong2021global} and the 2019 Kiney Segmentation Challenge (KiTS) dataset \cite{heller2019kits19}. For the LA dataset, we randomly divide the 100 scans into 80 training samples and 20 testing samples. For the KiTS dataset, we randomly split the total 210 images into 190 and 20 samples for training and testing, respectively.
\noindent
\textbf{Implementation Details} We implement our method on an NVIDIA A30 Tensor Core GPU using the PyTorch framework. 
We utilize the Adam optimizer with a fixed learning rate of 1e-4 for the registration module. 
We leverage MT \cite{tarvainen2017mean} as the baseline framework of our SPSS, where the teacher model is employed to generate pseudo labels and updated by an exponential moving average (EMA) of the student model. 
The student model is trained using the SGD optimizer for 6000 iterations, where the initial learning rate is 0.01 and gradually decays by 0.1 every 2500 iterations.
Following PLN \cite{li2022pln}, we randomly select the middle slice in a volume for image registration. 
The warm-up function is $\xi(t)=min(0.1 \times exp[-5(1-t/t_{max})^2],1)$. 
The age parameter is set as $\lambda = \alpha \cdot \delta$, which is an initial value $\alpha = 0.1$ growing with a factor $\delta = 1.01$. Please refer to the supplementary materials for the investigation of these hyperparameters.

\begin{table*}[!t]
\caption{Comparison with state-of-the-art methods on the LA dataset}
\label{LA_result_table}
\resizebox{\textwidth}{!}{
\begin{tabular}{cc|cc|cccc}
\hline
\multicolumn{2}{c|}{\multirow{2}{*}{Method}}                    & \multicolumn{2}{c|}{Scans Used}             & \multicolumn{4}{c}{Metrics}                                      \\ \cline{3-8} 
\multicolumn{2}{c|}{}                                           & \multicolumn{1}{c|}{L/U}   & Labeled Slices & DSC(\%)       & Jaccard(\%)    & ASD(voxel)    & HD(voxel)      \\ \hline
\multicolumn{1}{c|}{\multirow{7}{*}{Barely-supervised}} & MT \cite{tarvainen2017mean}    & \multicolumn{1}{c|}{16/64} & 16             & 59.80          & 42.97          & 13.44         & 34.55          \\
\multicolumn{1}{c|}{}                                   & UA-MT \cite{yu2019uncertainty} & \multicolumn{1}{c|}{16/64} & 16             & 63.11          & 46.78          & 17.01         & 44.15          \\
\multicolumn{1}{c|}{}                                   & CPS \cite{chen2021semi}   & \multicolumn{1}{c|}{16/64} & 16             & 59.19          & 42.79          & 20.00         & 50.83          \\
\multicolumn{1}{c|}{}                                   & FixMatch \cite{sohn2020fixmatch}   & \multicolumn{1}{c|}{16/64} & 16             & 61.75          & 44.94          & 11.98          & 29.87          \\
\multicolumn{1}{c|}{}                                   & UniMatch \cite{yang2023revisiting}   & \multicolumn{1}{c|}{16/64} & 16             & 60.74          & 44.13          & 13.21          & 31.30          \\
\multicolumn{1}{c|}{}                                   & PLN \cite{li2022pln}   & \multicolumn{1}{c|}{16/64} & 16             & 84.01          & 72.76          & 5.62          & 22.07          \\
\multicolumn{1}{c|}{}                                   & SPSS (ours)  & \multicolumn{1}{c|}{16/64} & 16             & \textbf{86.19} & \textbf{75.89} & \textbf{3.49} & \textbf{13.54} \\ \hline
\multicolumn{1}{c|}{Semi-supervised}                    & MT \cite{tarvainen2017mean}    & \multicolumn{1}{c|}{16/64} & 1280           & 88.12          & 79.03          & 2.65          & 10.92          \\ \hline
\end{tabular}
}
\end{table*}

\begin{table*}[!t]
\caption{Comparison with state-of-the-art methods on the KiTS dataset}
\label{KiTS_result_table}
\resizebox{\textwidth}{!}{
\begin{tabular}{cc|cc|cccc}
\hline
\multicolumn{2}{c|}{\multirow{2}{*}{Method}}                    & \multicolumn{2}{c|}{Scans Used}              & \multicolumn{4}{c}{Metrics}                        \\ \cline{3-8} 
\multicolumn{2}{c|}{}                                           & \multicolumn{1}{c|}{L/U}    & Labeled Slices & DSC(\%)   & Jaccard(\%) & ASD(voxel) & HD(voxel)  \\ \hline
\multicolumn{1}{c|}{\multirow{7}{*}{Barely-supervised}} & MT \cite{tarvainen2017mean}    & \multicolumn{1}{c|}{38/152} & 38             & 73.58          & 58.21           & 18.76          & 58.21          \\
\multicolumn{1}{c|}{}                                   & UA-MT \cite{yu2019uncertainty} & \multicolumn{1}{c|}{38/152} & 38             & 76.24          & 61.61           & 3.02          & 8.00          \\
\multicolumn{1}{c|}{}                                   & CPS \cite{chen2021semi}   & \multicolumn{1}{c|}{38/152} & 38             & 76.02          & 63.32           & 4.96          & 20.83          \\
\multicolumn{1}{c|}{}                                   & FixMatch \cite{sohn2020fixmatch}   & \multicolumn{1}{c|}{38/152} & 38             & 74.32          & 61.10           & \textbf{1.47}          & 8.83          \\
\multicolumn{1}{c|}{}                                   & UniMatch \cite{yang2023revisiting}   & \multicolumn{1}{c|}{38/152} & 38             & 74.81          & 61.55           & 1.58          & 8.85          \\
\multicolumn{1}{c|}{}                                   & PLN \cite{li2022pln}   & \multicolumn{1}{c|}{38/152} & 38             & 88.97          & 81.02           & 1.83          & \textbf{5.98}          \\
\multicolumn{1}{c|}{}                                   & SPSS (ours)  & \multicolumn{1}{c|}{38/152} & 38             & \textbf{90.22} & \textbf{83.55}  & 1.72 & 8.48 \\ \hline
\multicolumn{1}{c|}{Semi-supervised}                    & MT \cite{tarvainen2017mean}    & \multicolumn{1}{c|}{38/152} & 2432           & 92.56      & 87.67       & 1.70       & 6.26       \\ \hline
\end{tabular}
}
\end{table*}

\subsection{Comparison with SOTA}
We compare SPSS with SOTA semi-supervised and barely-supervised methods: MT \cite{tarvainen2017mean}, UA-MT \cite {yu2019uncertainty}, CPS \cite{chen2021semi}, FixMatch \cite{sohn2020fixmatch}, UniMatch \cite{yang2023revisiting}, PLN \cite{li2022pln}.

\subsubsection{Results on LA} In Table \ref{LA_result_table} and Fig. \ref{LA_result_image}, we present the quantitative and qualitative results on the LA dataset. 
Specifically, the SSL methods show inferior performance for barely-supervised left atrial segmentation.
This result suggests that the SSL approaches cannot handle the barely-supervised learning problem.
When compared to the barely-supervised SOTA, i.e., PLN, our method achieves an improvement of 2.18\% in terms of DSC. 
Remarkably, our SPSS, utilizing only 16 labeled slices, shows comparable performance against the SSL method that employs 1280 labeled slices.

\subsubsection{Results on KiTS} 
\begin{figure*}[!t]
    \centering
    \includegraphics[width=12cm]{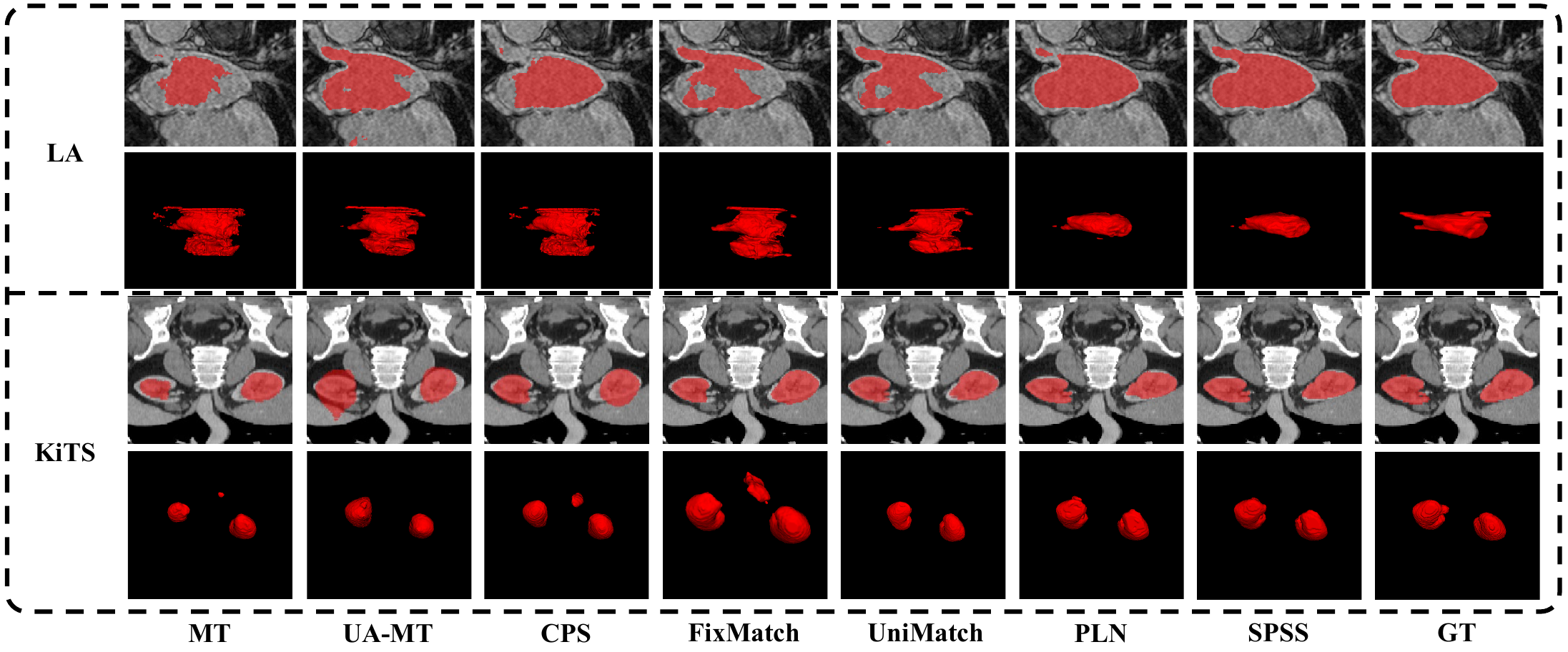}
    \caption{Qualitative results on the LA dataset and KiTS dataset.}
    \label{LA_result_image}
\end{figure*}

We further evaluate our SPSS on the KiTS dataset. Table \ref{KiTS_result_table} and Fig. \ref{LA_result_image} report the quantitative and qualitative results. 
Quantitatively, SPSS obtains the best performance among all the compared methods, with 90.22\% DSC, 83.55\% Jaccard, 1.72 ASD, and 8.48 HD. 
Surprisingly, SPSS, trained using only 38 labeled slices, attains performance close to MT trained with 1280 labeled slices.
Similar to the situation on the LA dataset, our SPSS outperforms the SSL methods by a large margin in the barely-supervised scenario. For example, SPSS surpasses UniMatch \cite{yang2023revisiting} by 15.41\% DSC under 38 labeled slices.
Furthermore, compared with PLN \cite{li2022pln}, our method obtains a DSC gain of 1.22\%.
Qualitatively, SPSS shows more accurate and smoother segmentation results, especially on the segmentation boundaries.
These results further demonstrate the effectiveness and advantages of SPSS for barely-supervised medical image segmentation.

\subsection{Ablation Study}
Table \ref{ablation} reports the results on the LA dataset with 20\% barely-annotated labeled data. 
Following PLN \cite{li2022pln}, we employ the registration-segmentation paradigm built upon MT \cite{tarvainen2017mean} as the baseline.
The results suggest that SU and SC contribute significantly to the performance improvement, which can be attributed to two main reasons: 1) SU allows for the selection of voxels with high-quality pseudo labels, and 2) SC enhances the separability between inter-class semantics in the feature space, further implicitly improving the quality of pseudo-labels. 


\begin{table}[!t]
\centering
\caption{Ablation study on the LA dataset with 20\% barely-annotated labeled data.}
\begin{tabular}{c|cccc}
\hline
Method   & DSC(\%) & Jaccard(\%) & ASD(voxel) & HD(voxel) \\ \hline
Baseline & 84.01   & 72.76       & 5.62       & 22.07     \\
Baseline + SU       & 84.73   & 73.71       & 4.05       & 14.64     \\
Baseline + SC       & 84.45   & 73.32       & 5.13       & 20.01     \\ \hline
Baseline + SU + SC (SPSS)     & 86.19   & 75.89       & 3.49       & 13.54     \\ \hline
\end{tabular}
\label{ablation}
\end{table}

\section{Conclusion}
We propose a novel framework, called SPSS, for barely-supervised medical image segmentation.
We pinpoint that the limitation of the registration-segmentation paradigm lies in the noisy pseudo labels generated by image registration.
Motivated by this, our main idea is to improve the quality of pseudo labels in both the image and feature spaces guided by self-paced sample selection.
Extensive experiments on two public datasets, including left atrial and kidney segmentation tasks, demonstrate the effectiveness of our SPSS and suggest its capability to achieve state-of-the-art performance.

\section*{Acknowledgment}
This research was supported by the National Natural Science Foundation of China (No.62076059) and the Science and Technology Joint Project of Liaoning Province (2023JH2/101700367, ZX20240193).

\section*{Disclosure of Interests}
The authors declare that they have no conflict of interest.

%
%
%
\bibliographystyle{splncs04}
\bibliography{ref.bib}
\end{document}